\documentclass[10pt,twocolumn,letterpaper]{article}

\usepackage{iccv}
\usepackage{times}
\usepackage{epsfig}
\usepackage{graphicx}
\usepackage{amsmath}
\usepackage{amssymb}

\usepackage{booktabs, multirow}
\usepackage[linesnumbered,ruled,vlined]{algorithm2e}

\usepackage{comment}
\usepackage{ dsfont }
\usepackage{cuted}
\usepackage{capt-of}
\usepackage{verbatim} 

\usepackage[pagebackref=true,breaklinks=true,letterpaper=true,colorlinks,bookmarks=false]{hyperref}

\iccvfinalcopy 

\def\iccvPaperID{4652} 

\ificcvfinal\fi
\begin{document}

\title{Synthesising 3D Facial Motion from ``In-the-Wild" Speech}

\author{Panagiotis Tzirakis$^1$, Athanasios Papaioannou$^{1,2}$, Alexander Lattas$^1$, Michail Tarasiou$^1$, \\ Bj{\"o}rn Schuller$^{1,3}$, Stefanos Zafeiriou$^{1,4}$\\
$^1$ Department of Computing, Imperial College London, UK\\
$^2$ Great Ormond Street Institute of Child Health, University College London, UK\\
$^3$ ZD.B. Chair of Embedded Intelligence for Health Care and Wellbeing, University of Augsburg, Germany \\
$^4$ Center for Machine Vision and Signal Analysis, University of
Oulu, Finland
}

\maketitle
\begin{strip}\centering
\includegraphics[width=1\textwidth]{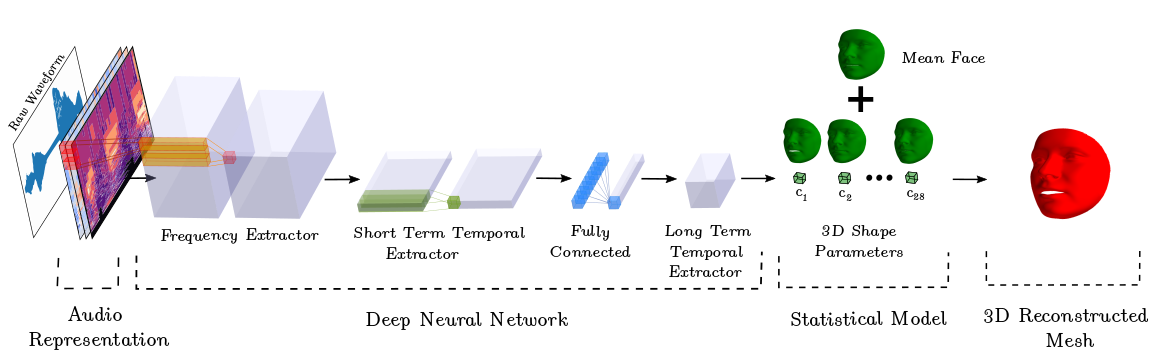}
\captionof{figure}{\small{Extraction of 3D facial mesh from raw in-the-wild speech signal.}}
\label{fig:Blendshapes2}
\end{strip}

\begin{abstract}
Synthesising 3D facial motion from speech is a crucial problem manifesting in a multitude of applications such as computer games and movies. Recently proposed methods tackle this problem in controlled conditions of speech. In this paper, we introduce the first methodology for 3D facial motion synthesis from speech captured in arbitrary recording conditions ("in-the-wild") and independent of the speaker. For our purposes, we captured 4D sequences of people uttering $500$ words, contained in the Lip Reading Words (LRW) a publicly available large-scale in-the-wild dataset, and built a set of 3D blendshapes appropriate for speech. We correlate the 3D shape parameters of the speech blendshapes to the LRW audio samples by means of a novel time-warping technique, named Deep Canonical Attentional Warping (DCAW), that can simultaneously learn hierarchical non-linear representations and a warping path in an end-to-end manner.
We thoroughly evaluate our proposed methods, and show the ability of a deep learning model to synthesise 3D facial motion in handling different speakers and continuous speech signals in uncontrolled conditions. 
\end{abstract}

\section{Introduction}

Synthesis of 3D talking faces is very crucial to many applications including but not limited to computer games, movies post-production (e.\,g., dubbing),
talking faces in virtual reality applications, etc. Currently, the highest quality 3D face synthesis is performed by using facial capture rigs which make use of markers or sensors. Recently, machine learning methods~\cite{edwards2016jali, Karras3d, phamend}, and in particular deep neural networks, have been used in order to train systems that reconstruct the 3D facial geometry of talking faces directly from audio sources. Nevertheless, till now the proposed methods are person or rig-specific \cite{visemenet} \footnote{In \cite{Karras3d} even though the method was trained on a single person, the authors claim that they used the method on other people and the results were meaningful. The above contradicts the current practices in machine learning which require large and diverse sets for good generalisation. Because we could not reproduce their experiments, we contacted the authors but received no reply.}, hence do not generalise to arbitrary audio sequences. In this paper, we present the first methodology for estimation of the 3D facial motion related to speech directly from raw audio stream. 

Estimating the 3D facial motion directly from raw audio samples captured in arbitrary recording conditions is an ill-pose problem since a great number of people uttering a large amount of diverse words need to be captured in 4D (i.e. 3D geometry in time). Even though many efforts have been performed towards collecting 4D expressive faces~\cite{cheng20184dfab, yin20063d,yin2008high} there is a lack of datasets with talking 4D faces (i.e., 3D speech in time). One such dataset has been proposed by Marshall et al.~\cite{marshall20154d}, which captured four people in dyadic interaction (17\,mins in total). A limited number of words have been captured in~\cite{cheng20184dfab} for biometric application, nevertheless the data are not publicly available. Hence, it is very difficult to train a generic method for 3D facial motion reconstruction from audio streams. The most closely related work to ours is by Phamend et al.~\cite{phamend}, which used a statistical blendshape model, trained on facial expressions. As we show, these blendshapes cannot model adequately 3D facial motion related to speech.

\begin{figure}\centering
\includegraphics[width=8.4cm,height=3.3cm]{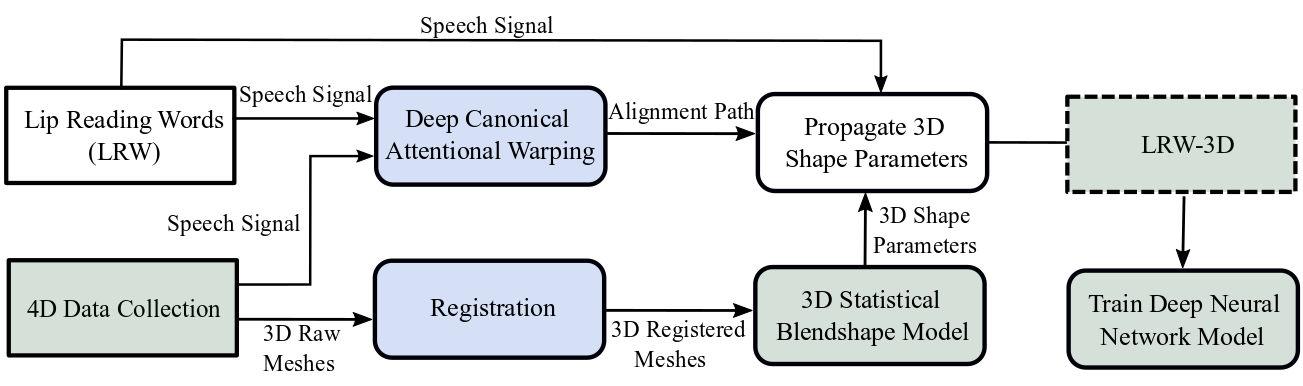}
\label{fig:process}
\caption{Pipeline of our approach. The proposed methods are represented in light blue, whereas the dataset and models, that we provide publicly available, are shown in light green.}
\end{figure}

In this paper, we make the first, to the best of our knowledge, comprehensive effort to estimate 3D facial motion from arbitrary audio streams. To this end, we first capture 4D sequences of people uttering $500$ words, contained in an in-the-wild dataset named Lip Reading Words (LRW). After registering the 3D meshes with an adaptive template approach, we learn 3D blendshapes for the speech, which we make publicly available.  Each 3D mesh can be parameterised as a set of 3D shape parameters through these 3D blendshapes. Employing a novel time-warping algorithm, named Deep Canonical Attentional Warping (DCAW), we align the speech, that we have 3D ground-truth on, 
with the corresponding "in-the-wild" speech signals in LRW, and propagate the 3D shape parameters, creating the "in-the-wild" LRW-3D dataset. We train a deep learning model on this dataset and show the ability of the model to predict 3D face motion for speech captured under uncontrolled conditions. Fig.~\ref{fig:process} shows the pipeline of our approach. In summary, the contributions of this work are the following: 

\begin{itemize} 
\item[1.] We collect a 4D dataset of people uttering $500$ words and learn the first statistical blendshape model for speech which we provide publicly available. 
\item[2.] In order to train accurate blendshapes, we propose an adaptive shape template method to accelerate the convergence of registration algorithms and achieve a better final shape correspondence.
\item[3.] We propose Deep Canonical Attentional Warping (DCAW), a method which learns hierarchical non-linear representations and temporal alignment of two audio signals in an end-to-end manner. Using DCAW we create LRW-3D, and we make publicly available the aligned 3D shape parameters.
\item[4.] Finally, we train a speech to 3D deep facial motion model that can operate in nearly real-time, and independently of the speaker in uncontrolled conditions of speech.
\end{itemize}


\section{Related Work}
\label{related_work}

There are several traditional approaches~\cite{fan2015photo, katsamanis2009face,  mattheyses2015audiovisual, sako2000hmm, ohman1999using, salvi2009synface, wang2007assembling, xie2007realistic} that exploit audio signal for 2D or 3D facial animation. More recent approaches utilise deep neural networks (DNN) for this task. These can be categorised in two main groups: linguistic-driven and audio-driven approaches. 

\textbf{Linguistic-driven approaches}. Language-based methods take advantage of the mapping between phonemes and their visual counterpart visemes. For example, Edwards et al.~\cite{edwards2016jali} proposed the JAw and LIp (JALI) model, a two-dimensional space that represents the jaw and lip movements of a facial animation based on psycholinguistic considerations. The main disadvantage of their study is the need for the speech signal, its text transcript and their alignment to create the facial animation. In another study, Taylor et al.~\cite{taylor2012dynamic} first proposed generating dynamic units for visual speech for realistic visual speech animation. In a more recent study, the authors~\cite{taylor2017deep} initially  transcribe the speech signal to phoneme labels, which are then fed to a deep fully connected network to predict person-specific shape and appearance parameters obtained by Active Appearance Model (AAM). A main limitation of this work is the need for speech to phoneme labels conversion.  

\textbf{Audio-driven approaches}. Audio-based methods drive facial animations using only audio cues. Zhou et al.~\cite{visemenet} utilises audio features for an automatic and near real-time animation by driving a JALI face-rig. The authors propose VisemeNet, a three stage deep learning model based on Long Short-Term Memory (LSTM) networks that are fed with audio features. The first two stages comprises of extracting phonemes and landmarks, while the third extracts visemes. In a different study, Karras et al.~\cite{Karras3d} proposed a deep convolutional neural network for 3D facial animation using auto-correlation audio features. The network models unknown variation of the audio cues by learning a small dimensional vector. Its output is the entire face and is trained using a three-way loss function that ensures temporal stability and correct responses under animation. The main limitation of their method is that the trained network is person-specific and as such it cannot generalise well to different speakers. Pham et al.~\cite{phamend} used the same network architecture as Karras et al. but with melspectrograms as input. Their model was trained to predict the rotation and expression blending parameters extracted by a 3D face tracker. Suwajanakorn et al.~\cite{suwajanakorn2017synthesizing} used audio features to synthesize videos of Obama focusing on the mouth region and taking the rest of the head and torso from stock footage. However, their method is person-specific with tens of hours of audio signal, and requires heavy hand-engineered work. 

In comparison with the aforementioned methods, our approach is able to provide 3D face motion estimation not only in uncontrolled speech conditions but also is speaker-independent, i.\,e., independent of the speaker. On top of that, we are the first to propose a statistical blendshape model appropriate for speech.


\section{Building Speech Blendshapes Model}
\label{sec_speech_blendshapes}

In this section, we describe the pipeline for building the speech blendshapes model.
The process consists of: (a) collecting a dataset of 3D meshes with various people uttering a set of words (Sec.~\ref{data_acquisition}), (b) registering all the meshes to a common template (Sec.~\ref{registration}), and (c) building a statistical model on the registered meshes (Sec.~\ref{speech_blendshapes}).

\subsection{Data Acquisition}
\label{data_acquisition}

To construct a set of blendshapes appropriate for speech we needed to capture 4D sequences (i.\,e. 3D geometry in time) of people talking. The choice of the utterance spoken by our participants is driven by our goal to train a model that can operate under unconstrained audio conditions (in-the-wild), and is speaker independent. To this purpose, we utilise the $500$ words contained in the publicly available Lip Reading Words (LRW) in-the-wild dataset~\cite{chung2016lip} which contains almost $1000$ videos per word, summing to approximately $450,000$ videos. These videos were captured from TV broadcasts (e.\,g. news or interviews) in uncontrolled environments and contain $1000$ speakers, making it an excellent fit for our purposes. 

We used the DI4D dynamic system\footnote{http://www.di3d.com} to capture and build 4D faces. This system consists of six cameras (two pairs of stereo cameras and one pair of texture cameras, $30$FPS, $1200\times1600$). Before every recording, a calibration was necessary and was performed by utilising a $10\times10$, 20\,mm checkerboard. Two 4-lamp fluorescent lights were placed on each side to provide consistent and uniform lights. One microphone is used to capture the audio signal in $44,1$\,kHz sampling rate.

We captured 2 native and 2 non-native English speakers reading out the $500$ words from the aforementioned dataset. We should note that the choice of the non-native speakers is to add extra variability to the mouth region when we develop our blendshapes. In total, the whole process took 20\,min for each participant, and, approximately, $20,000$ 3D meshes were acquired for each individual (equivalent of $660$\,s of recording).

\subsection{Registration}
\label{registration}

\textbf{Automatic Annotation.}
Before processing our 3D captured meshes, we need to re-parameterise them such that all the meshes have the same number of vertices joined into a common triangulation. Two approaches exist that perform such task and they differ on the space the registration is performed. The first method performs the dense registration in the 2D space, namely the UV-space~\cite{patel20093d, cosker2011facs}. The second method registers directly (i.\,e. in the 3D space) the mesh and the template~\cite{amberg2007optimal, myronenko2010point}.

In this paper, we follow the latter approach as UV-based correspondence approaches introduce non-linearities into the process and require an extra step for rasterizing the UV image~\cite{booth2018large}. To this end, we perform the registration in the 3D space between a neutral 3D mesh and the mean shape of Large Scale Facial Model (LSFM)~\cite{booth2018large}. The registration is performed by utilising Non-rigid Iterative Closest Point (NICP), which has as a prerequisite the template and the mesh to be close in terms of euclidean distance.  

The high number of 3D meshes do not allow us to manually annotate 3D facial landmarks. To automate the process we utilise a sparse alignment method proposed by Booth et al.~\cite{booth20163d} that can automatically compute sparse annotations on the 3D meshes. More specifically, for each mesh we apply the face detection and alignment framework proposed by Zhang et al.~\cite{zhang2016joint} on its corresponding 2D texture image, where the correspondence to the 3D coorditanes are known,  to robustly locate a set of $68$ sparse annotations (landmarks)~\cite{zafeiriou20183d} in the 2D space. Exploiting the  correspondence between the texture image and the 3D mesh, we get the corresponding 3D positions of the landmarks.

\textbf{Adaptive Template and Dense Registration.}
The majority of NICP shape registration methods use the same 3D template shape to deform all 3D meshes of a dataset~\cite{booth2018large}. Even though this approach can be sufficient for large datasets with neutral shapes, it is not the case when there is a large variation in terms of expressions and lip movements in each mesh.
In our captured 3D meshes the position and shape of the mouth change frequently so if a single template shape is used important parts of it would not be close to the corresponding parts of 3D meshes. Hence, the registered results would have visible errors and inaccurate correspondences. 

\begin{figure*}
\centering
\includegraphics[width=\textwidth]{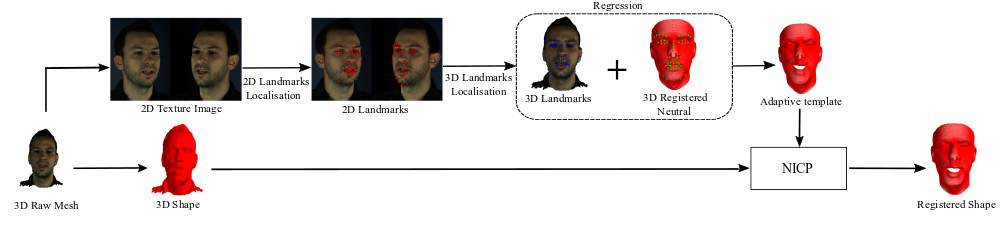}
\captionof{figure}{\small{Registration pipeline. The process starts by extracting 2D landmarks from the texture image, and then their corresponding position in the 3D mesh. By applying a regression between the 3D landmarks of the raw mesh and the neutral registered mesh we create an adaptive template. Using this template and the 3D shape of the raw mesh NICP can accurately register the raw mesh.}}
\label{fig:registration_pipeline}
\end{figure*}

To alleviate from this problem, we propose an adaptive template approach where for each 3D mesh in our dataset we adapt the original template using sparse shape information (i.\,e. $68$ landmarks). 

In particular, by leveraging the expression blendshapes created by Cheng et al.~\cite{cheng20184dfab}, we compute the 3D shape parameters for each mesh through linear regression between the landmarks of the neutral shape and the landmarks of the reconstructed instance as 

\begin{equation}
  \mathbf{c}_o =   \arg\min_{\mathbf{c}} || \boldsymbol l_{n} - \mathbf{A}(\mathbf{x}_{n}+\mathbf{U}_s\mathbf{c})||^2_F 
  \label{Eq:Min}
\end{equation}

where $\boldsymbol l_{n} \in \mathbb{R}^{3m}$ is a vector with $m$ landmarks of the neutral shape, $\mathbf{A} \in \mathbb{R}^{3m \times 3n}$ is an indicator matrix, indicating the position of the 3D landmarks on the reconstructed mesh, $\mathbf{x}_{n} \in \mathbb{R}^{3n}$ is the neutral registered shape, $\mathbf{U}_s \in \mathbb{R}^{3n \times q}$ is a matrix with the $q$ blend shapes and $\mathbf{c} \in \mathbb{R}^{q}$ is the 3D shape parameters vector.

After the calculation of the 3D shape parameters $\mathbf{c}_o$ that minimise the expression in Eq.~\ref{Eq:Min}, we can adapt the neutral template to the current mesh: $\mathbf{x}_{adapt} = \mathbf{x}_n + \mathbf{U}_s\mathbf{c}_o$. Finally, NICP can be performed between the adaptive template and the mesh. We should note that even though the blendshapes ($\mathbf{U}_s$) describe various expressions and not speech, they give a good prior for our template. The registration process is illustrated Fig.~\ref{fig:registration_pipeline}.

\subsection{Creating Speech Blendshapes}
\label{speech_blendshapes}

Our final step is to build a set blendshapes appropriate for speech. We start by subtracting the neutral mesh from each registered mesh in the sequence, creating vectors of differences, i.\,e., $\mathbf{d} = \mathbf{x}_{n} - \mathbf{x}_{adapt} \in \mathbb{R}^{3n}$. After we stack these vectors into a matrix $\mathbf{D} = [\mathbf{d}_1, \dots, \mathbf{d}_T] \in \mathbb{R}^{3n \times T}$, we apply Principal Component Analysis (PCA) to identify  the deformation components $\mathbf{U}_b$. We keep $28$ blendshapes corresponding to $99.9\%$ of the total variance in the sequence. Hence, a new shape instance can be generated: $\mathbf{x}_{new} = \mathbf{x}_n + \mathbf{U}_b \boldsymbol \lambda$, where $\boldsymbol \lambda$ are the 3D shape parameters of our model. Finally, for each mesh in our sequence, we compute the 3D shape parameters that constitute our ground truth.

\section{Lip Reading Words in 3D (LRW-3D)}
\label{dcaw_3d_propagation}

With the construction of our speech-driven statistical model, we can now use the speech signal of our participant to propagate the 3D shape parameters to the speech signals of the LRW dataset.
Our process starts by segmenting our participants' speech signal to the $500$ words, which is accomplished in a semi-automatic manner. More particularly, we first utilise the approach proposed by Elsner et al.~\cite{elsner2017speech} to segment our speech signal. This step is crucial to our pipeline as a single faulty segmentation can result in almost a thousand false samples in our dataset, which can lead to ill-generalisable models. To this purpose, we listened each segment, and, when required, we manually adjusted them. 


\subsection{Deep Canonical Attentional Warping (DCAW)}
To accurately propagate the 3D shape parameters of our participant to the LRW dataset, we need to eliminate any temporal variations arising in the data.
Hence, we compute a temporal alignment between the signal of each word uttered by our participant and the corresponding signals of the LRW. To compute this alignment, we propose \textit{Deep Canonical Attentional Warping (DCAW)}, a novel method that can maximally correlate two data sequences (or views) and find a temporal alignment in an \textit{end-to-end} manner. We leverage deep recurrent convolutional neural networks to spatially transform the raw speech signals, and utilise attention mechanism for the alignment.

The \textit{attentional warping} is performed by computing attention weights between each feature frame of the one view (source) with all the features from the other view (target). We should point out that our data are monotonic and as such we utilise a monotonic attention mechanism~\cite{raffel2017online}. Mathematically, given two views $\mathbf{X}_k \in \mathbb{R}^{d_k \times T_k}$, and their features $\textbf{h}_i^k$, $i = \{1, 2, ..., T_k\}$, where $d_k$ and $T_k$ are the dimensionality and length of view $k \in \{1, 2\}$, respectively, the attention between each target feature and the source ones is computed as follows:

\begin{align}
\label{eq:attention1}
    \alpha_{\textbf{h}_i^1, \textbf{h}_j^2} = \frac{\exp(\operatorname{score}(\textbf{h}_i^1, \textbf{h}_j^2))}{\sum_{k=1}^{T_2} \exp(\operatorname{score}(\textbf{h}_i^1, \textbf{h}_k^2))},
\end{align}

where $\exp$ is the exponential function, and $\operatorname{score}$ can be any attention function, such as Bahdanau~\cite{bahdanau2014neural} or Luong~\cite{luong2015effective}. For our purposes, we use the Bahdanau score: 

\begin{align}
\label{eq:attention2}
    \operatorname{score}(\mathbf{h}_i^1, \mathbf{h}_j^2) = \mathbf{v}_i^T \tanh(\mathbf{W}_t\mathbf{h}_i^1 + \mathbf{W}_s\mathbf{h}_j^2),
\end{align}

where $\mathbf{v}_i^T$, $\mathbf{W}_t$ and $\mathbf{W}_s$ are learnable parameters of the model.

The outcome of Eq.~\ref{eq:attention1}, is the attentional matrix $\mathbf{A} \in \mathbb{R}^{T_1 \times T_2}$ that includes the weights between the two views.
We use this matrix to warp the target features with the source ones by multiplying it with the source features, i.\,e., $\mathbf{H}^1\mathbf{A}$, where $\mathbf{H}^1 \in \mathbb{R}^{d \times T_1}$ is the feature matrix containing all the features of the first view.

The alignment between the two views is found such that the features between the views are maximally correlated. This optimisation is formulated as a least-square problem:


\begin{equation}
\begin{array}{rl}
\arg\min_{\theta_1, \theta_2} &|| f_1(\mathbf{X}_1;\theta_1) - f_2(\mathbf{X}_2;\theta_2) ||_F^2 
\\ \\
\mbox{subject to}\, &f_1(\mathbf{X}_1;\theta_1)f_1(\mathbf{X}_1;\theta_1)^T = \mathbf{I} \\
&f_2(\mathbf{X}_2;\theta_2)f_2(\mathbf{X}_2;\theta_2)^T = \mathbf{I}\\ 
&f_1(\mathbf{X}_1;\theta_1)f_2(\mathbf{X}_2;\theta_2)^T = \mathbf{D} \\
&f_1(\mathbf{X}_1;\theta_1)\mathbf{1} = f_2(\mathbf{X}_2;\theta_2)\mathbf{1} = \mathbf{0},
\end{array}
\label{Projection_matrix}
\end{equation}
\\
where $f_k(\mathbf{X}_k;\theta_k)$ with $k \in \{1, 2\}$ represents the output of the two neural networks with parameters $\theta_k$ and input $\mathbf{X}_k$, respectively, $\mathbf{D}$ is a diagonal matrix, and $\mathbf{1}$ is an appropriate dimensionality vector of all ones. 

We find the optimal parameters for each network with the use of backpropagation. Our problem is a variant of Deep Canonical Correlation Analysis (DCCA)~\cite{andrew2013deep}, and as such the optimal objective value can be computed as the sum of the $k$ largest singular values of $\mathbf{K}_{DCAW} = \mathbf{\Sigma}_{11}^{-1/2} \mathbf{\Sigma}_{12} \mathbf{\Sigma}_{22}^{-1/2}$, where  $\mathbf{\Sigma}_{ij} = \frac{1}{T_2 - 1} f_i(\mathbf{X}_i;\theta_i) \mathbf{C}_T f_j(\mathbf{X}_j;\theta_j)^T$ and $\mathbf{C}_T = \mathbf{I} - \frac{1}{T_2} \mathbf{1} \mathbf{1}^T$ is the centering matrix. The optimal objective is found by maximising the nuclear norm $||\mathbf{K}_{DCAW} ||_{*} = tr(\sqrt{\mathbf{K}\mathbf{K}^T})$, i.\,e.,

\begin{align}
\label{eq:on2}
    \arg\max_{\theta_1, \theta_2} ||\mathbf{K}_{DCAW} ||_{*}
\end{align}

We use gradient ascent to optimise Eq.~\ref{eq:on2}. Since the gradient cannot be computed analytically we use the subgradient~\cite{bach2008consistency} by computing the singular value decomposition of $\mathbf{K}_{DCAW} = \mathbf{U} \mathbf{S} \mathbf{V}^T$, then the subgradient for the last layer of the network can be defined as follows

\begin{align}
\label{eq:attenti}
    \mathbf{L}_+ &= \mathbf{\Sigma}_{11}^{-1/2} \mathbf{U} \mathbf{V}^T  \mathbf{\Sigma}_{22}^{-1/2} f_2(\mathbf{X}_2;\theta_2)\mathbf{C}_T \\
    \mathbf{L}_- &= \mathbf{\Sigma}_{11}^{-1/2} \mathbf{U} \mathbf{S} \mathbf{U}^T  \mathbf{\Sigma}_{11}^{-1/2} f_1(\mathbf{X}_1;\theta_1)\mathbf{C}_T \\
    &\frac{\vartheta ||\mathbf{K}_{DCAW} ||_{*}}{\vartheta f_1(\mathbf{X}_1;\theta_1)} = \frac{1}{T_2 - 1} ( \mathbf{L}_+ - \mathbf{L}_-).
\end{align}

Finally, we should point out that DCAW can be extended to handle multiple data sequences by utilising an objective similar to Multi-set CCA, i.\,e., 

\begin{align}
  \sum_{i,j} ||K_{DCAW}^{i,j}||_*.
\end{align}

\subsection{Word Alignment}


We can now use DCAW to compute an alignment path between the speech signal of words uttered by our participant and the corresponding signals of the LRW dataset. To this end, we train a recurrent convolutional neural network for each word (i.\,e. $500$ networks - see Sec.~\ref{training} for topology) by fixing one of the views to our participant's speech signal and the other to the corresponding speech signals of LRW, and get the alignment path between them.

Utilising the alignment path of the speech signals, we propagate the 3D shape parameters computed for our participant to the LRW dataset. 
In the case where an audio frame of the LRW is aligned to multiple frames from our dataset, the mean of the 3D shape parameters of these frames is computed and assigned as the ground truth of that frame. If the opposite holds, namely, an audio frame from our speech signal is aligned to multiple audio frames of the LRW, then the same ground truth is assigned to all LRW audio frames.

By transferring our 3D shape parameters to the LRW dataset, we create a large word-level dataset in-the-wild for 3D dense shape estimation from speech. This allows us to train models that can generalise to every speaker, and at the same time to in-the-wild speech signals.

\section{Training}
\label{training}

Three aspects are relevant to our training: (i) the input representation, (ii) the network topology, and (iii) the objective function utilised for training and evaluating the model. We describe in detail in the rest of the section.

\textbf{Input Representation.}
All audio signals have sampling rate at $44.1$\,kHz, and after we remove the DC offset, we normalise its volume to $0$\,dB, namely, using the full $[-1, 1]$ range. No other pre-processing step takes place. 

We use mel-spectograms as our input represenation of the audio signal. This representation is appropriate for our task because it is derived by approximating the frequencies perceived by the human cochlea~\cite{rabiner2011theory}. Thus, the information is similar to the perceived human hearing. 

For each visual frame, we derive mel-spectrograms in an audio window of length $400$\,ms so that we can take into account co-articulation effects in the signal. For each audio window, we compute $128$ mel-frequency parameters utilising a window of length $20$\,ms ($882$ samples) with $10$\,ms ($441$ samples) overlap. Hence, a 2D representation is formed of size $41 \times 128$. By calculating its first and second temporal derivatives, and place all three 2D representations in a different channel, we form a $41 \times 128 \times 3$ representation, which is the input to our convolutional recurrent neural network.

\textbf{Network Topology.}
Our deep neural network topology is inspired by Karras et al.~\cite{Karras3d}, and is comprised of four parts: (a) frequency extractor, where features are extracted vertically, namely, exploiting the frequency domain of the input representation, with kernel and stride size of 3 and 2, respectively, (b) short term temporal extractor, where features are extracted horizontally, namely, from the temporal domain of the extracted representation of the previous step, with kernel and stride size of 3 and 2, respectively, (c) a non-linear transformer, that non-linearly transform of the convolutional extracted features 128 dimensionality, and (d) a long term temporal extractor represented with a recurrent neural network of 1-layer LSTM cell of 128 dimensions, which captures the long term temporal dynamics in the data. The last layer is a fully connected that produces the 3D shape parameters of our model. The filter size for all convolution layers is set to 64.

\textbf{Objective Function.}
Most of the studies in the literature use as objective function the Mean Squared Error~(MSE). However, we propose to use an objective function that is based on the Concordance Correlation Coefficient ($\rho_c$), which is also used as our evaluation metric. The correlation coefficient evaluates the agreement level between the predictions and the ground truth by scaling their correlation coefficient with their mean square difference. More particularly, for each shape parameter $i$ we define the concordance loss $\mathcal{L}_{c}^i$ between the ground truth $\mathbf{x}$ and the prediction $\mathbf{y}$ as follows:

\begin{align}\nonumber
\mathcal{L}_{c}^i
&= 1 - \rho_c = 1 - \dfrac{2 \sigma_{xy}^2}{\sigma_x^2 + \sigma_y^2 + (\mu_x - \mu_y)^2}, \\
\label{eq:ccc_loss}
\end{align}
where $\mu_{x} = \mathbb{E}(\mathbf x)$, ${\mu_{y} = \mathbb{E}(\mathbf y)}, {\sigma_x^2 = \mbox{var}(\mathbf x)},  \sigma_y^2 = \mbox{var}(\mathbf y)$ and $\sigma^2_{xy} = \mbox{cov}(\mathbf x, \mathbf y)$. 

For our purposes we train our networks to simultaneously predict all $28$ 3D shape parameters. We should note that each shape parameter explains different variability percentage in our data, and hence contributes differently to the final 3D reconstructed mesh. We take this fact into account and we add as weight $w_i$ to each shape parameter concordance loss $\mathcal{L}_c^i$, the variability percentage the shape parameter represents. Our overall loss function is defined as $\mathcal{L} = \sum_{i=1}^{28} w_i \mathcal{L}_c^i$.


\section{Experiments}
\label{experiments}

We perform extensive evaluation of our proposed methods by: (i) comparing the DCAW with the current state-of-the-art method for maximally correlating two views and finding an alignment (Sec.~\ref{dcaw}), (ii) comparing the proposed speech blendshapes with the blendshapes of the FaceWarehouse (Sec.~\ref{blendshapes_comparison}), and (iii) validating our approach for 3D face motion from "in-the-wild" speech (Sec.~\ref{audio_experiments}).

\subsection{Experimental Setup}
\label{hyperpar}

The hyperparameters of the models are kept the same throughout all of the experiments. As our optimisation function, we use the Adam optimiser~\cite{adam} with $\beta_1 = 0.9$, $\beta_2 = 0.99$, and a initial learning rate of $5\times10^{-4}$, with batch size set to $50$. Finally, the initialisation of the feature extraction network was performed following He et al.~\cite{he2015delving}, whereas our recurrent network weights are initialised following Glorot based initialisation~\cite{glorot2010understanding}. We should note that zero padding was used to samples that do not match the maximum sequence length of the batch. We discard the zero-padded frames that do not belong to the sample by applying a mask.

\subsection{Deep Canonical Attentional Warping}
\label{dcaw}

We compare our proposed DCAW with the state-of-the-art method for representation learning and temporal warping, the Deep Canonical Temporal Warping (DCTW)~\cite{trigeorgis2018deep}. The performance of the two methods is evaluated on two datasets: (a) the MMI Facial Expression Dataset~\cite{pantic2005web}, which contains more than $2900$ videos of $75$ different subjects, each performing a particular combination of Action Units (i.\,e., facial muscle activations). We predict the AU12 and utilise the same approach and network architecture (with a recurrent network on top) as in~\cite{trigeorgis2018deep}. (b) We use the LRW dataset to perform template matching, where one of the views is a speech signal of a word and the other is a speech signal of a sentence containing, in a random location, the word. The methods need to accurately find the word in the sentence. For our purposes, we use $10$ randomly chosen words with $100$ samples. 

The performance measure is the alignment error introduced in~\cite{zhou2012generalized}. More particularly, given $m$ sequences, each algorithm infers a warping path, i.\,e., $P_{alg} = [p_{alg}^1, ..., p_{alg}^m] \in \mathbb{R}^{l_{alg} \times m}$, and the alignment error is computed with the ground truth path $P_{grd} = [p_{grd}^1, ..., p_{grd}^m] \in \mathbb{R}^{l_{grd} \times m}$ as follows:

\begin{equation}
    \text{Error} = \frac{dist(P_{grd}, P_{alg}) + dist(P_{alg}, P_{grd})}{l_{grd} + l_{alg}},
\end{equation}
where $dist(P_{grd}, P_{alg}) = \sum_{i=1}^{l_1} min_{j=1}^{l_2} ||p^{(i)}_1 - p^{(j)}_2||_2$.

Table~\ref{dcaw_result} depicts the results for both experiments. Our method outperforms DCTW in both of them, and for the LRW one we find the result to be statistically significant ($a < 0.05$).
These results validate our choice of using DCAW for aligning our participant's speech signal with the ones from the LRW dataset.

\begin{table}[ht]
\centering
\begin{tabular}{l|r|r}\toprule\midrule
Dataset & DCTW & DCAW \\\midrule
MMI & 0.59 & \textbf{0.61} \\
LRW & 0.64 & \textbf{0.72} \\
\end{tabular}
\caption{Results (wrt the mean alignment error) of the DCAW and DCTW methods on the MMI and LRW datasets.}
\label{dcaw_result}
\end{table}

\subsection{Blendshapes Comparison}
\label{blendshapes_comparison}

We compare quantitatively and qualitatively the proposed blendshapes with the FaceWarehouse ones, that were used by Pham et al.~\cite{phamend}, by testing the generalisation capacity of the blendshapes to represent unseen 3D facial meshes. 

\begin{figure}[!ht]
\centering
\includegraphics[width=8cm]{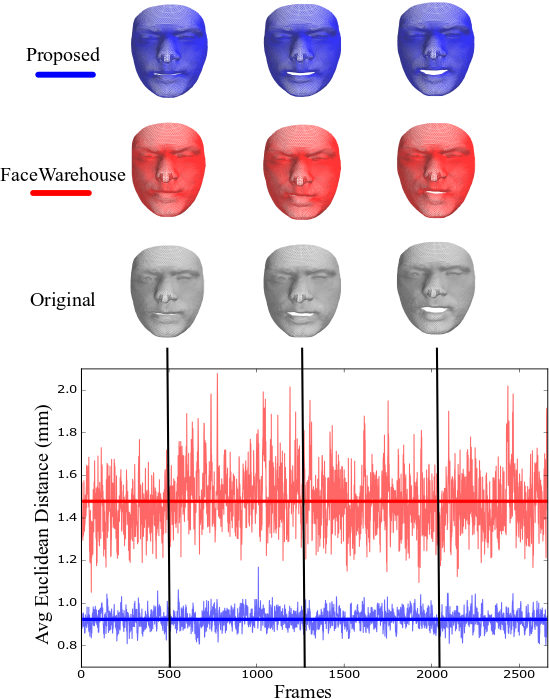}
\caption{Average Euclidean distance for a sequence of frames between the FaceWarehouse and the proposed blendshapes. - Best viewed in colour.}
\label{fig:Blendshapes_comparison}
\end{figure}

More particularly, we compute the error as the per-vertex Euclidean distance between every mesh of the test subject (i.\,e. not included in the training process) and its corresponding projection to the subspace defined by the blendshapes. An average value is computed over all vertices. For the whole sequence the mean error of the FareWarehouse is 1.29, and our proposed blendshapes is 0.87. Fig.~\ref{fig:Blendshapes_comparison} depicts the error in a sequence of $2,550$ frames. It is clear that the proposed blendshapes outperform the FaceWarehouse ones by a high margin. Finally, for qualitative purposes, we show three samples of the original 3D facial shapes and how they are reconstructed by the FaceWarehouse and the proposed blendshapes. 

\begin{figure}[!ht]
\centering
\includegraphics[height=8cm]{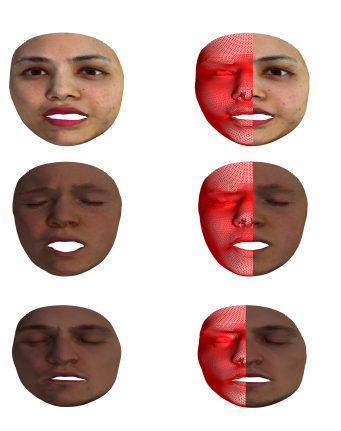}
\caption{Depicting the ground truth 3D shape parameters (left column) and the predictions (right column) for three individuals. - Best viewed in colour.}
\label{fig:qualitative_results}
\end{figure}

To further demonstrate the generalisation capability of our speech blendshapes, we show our model's prediction and the ground truth 3D shape parameters on three individuals. Fig.~\ref{fig:qualitative_results}, shows the results.

\begin{table*}[!htbp]
\centering
\begin{tabular}{l|l|c|c|c|c|c|c}\toprule\midrule
Dataset & Methodology & $\mu_{28}$ & $c_1$ & $c_2$ & $c_3$ & $c_4$ & $c_5$ \\\midrule
\multirow{2}{*}{LRW} & Speaker Ind. & $.621$~($.679$) & $.643$~($.712$) & $.607$~($.684$) & $.563$~($.596$) & $.652$~($.683$) & $.582$~($.641$) \\
& Word/Speaker Ind. & $.502$~($.554$) & $.536$~($.556$) & $.582$~($.618$) & $.557$~($.624$) & $.395$~($.405$) & $.411$~($.426$) \\ 
LRS & Continuous Speech & $.463$ & $.482$ & $.414$ & $.443$ & $.475$ & $.504$ \\
\end{tabular}
\caption{Results with respect to $\rho_c$ for the experiments: speaker independence, word/speaker independence, and sentences. The mean value of the estimation of the $28$ 3D parameters ($\mu_{28}$), and the first five 3D parameters is depicted. In parenthesis the results on the validation set.}
\label{word_results}
\end{table*}

\subsection{Audio Experiments}
\label{audio_experiments}

We start the validity of our method by performing two kind of experiments using the LRW-3D dataset: (i) \textit{speaker-independent}, where we measure the performance of our model on different speakers that pronounce the same words, and (ii) \textit{word- and speaker-independent}, where we evaluate our model on different speakers that pronounce different words. For both experiments our test set is comprised of one of the participant's audio signal with the corresponding 3D shape parameters, which are excluded from the training process. Finally, we measure the performance of our model on continuous speech signal.

\textbf{Speaker-Independent.}
In our first experiment we test the performance of the model when the same words are uttered by different individuals. To this end, we split the $1000$ speakers of the dataset to $900$ for training, and the rest $100$ for validation. Hence, our training set is comprised of $370,000$ samples, and our validation set of the rest $80,000$. Table~\ref{word_results} depicts the results of the mean, and the first five shape parameters in terms of the $\rho_c$ metric. The results indicate the validity of our method to generalise to every speaker and in uncontrolled conditions. 

\textbf{Word- and Speaker-Independent.}
Considering the high performance of the model for different speakers, in this experiment we test its performance for different words and speakers. Hence, we split our dataset to a training set that contains $450$ words and $900$ speakers, and the validation set contains the rest of the $50$ words and $100$ speakers. In total, the training set contains approximately $355,000$ samples, and the validation set approximately $95,000$ samples. We should point out that the words that comprise the validation set were chosen such that they contain the same phonemes as the ones in the training set. 

To improve the generalisation capacity and reduce overfitting of our model, we perform a random time-segmentation of our training samples. More particularly, each sample in the training batch is randomly segmented from its both ends but always keeping at least $50$\,\% of the frames of the original sample. This is particularly beneficial to our recurrent network architectures as now the temporal dynamics in the training set vary. 

Table~\ref{word_results} depicts the results of the mean, and the first five shape parameters in terms of the $\rho_c$ metric. The performance drops compared to the previous experiment as now the temporal information in the validation set is different from the training one. However, this does not limit the capacity of the model to be able to accurately reconstruct the 3D meshes. 

\textbf{Sentence-level Experiments.}
We also test the performance of our model in continuous speech signals. 
More specifically, we captured a native speaker (different than the previous experiments) pronouncing $50$ sentences (3 to 8\,sec long), taken from the Lip Reading Sentences (LRS) in-the-wild dataset~\cite{chung2017lip}, and extracted his 3D parameters. After the extraction of mel-spectrograms from the raw waveform, we feed them to the model and estimate its test performance in continuous speech signals. Table~\ref{word_results} depicts the results of the mean, and the first five shape parameters in terms of the $\rho_c$ metric. We observe that the performance of our model remains also high in this experiment. 
We should point out that our model is trained with short temporal dynamics ($10$ to $30$ frames long), and as such it cannot accurately predict 3D shape parameters for longer sequences such as sentences. We tackle this difficulty by splitting the speech signal of the sentences to sequences of $15$ frames long and feed them separately our model. We apply a temporal filter on the predictions of our model, to remove temporal discontinuities added by the LSTM. In the supplementary material videos are provided that show the effectiveness of our method.

\section{Conclusions}

We presented a methodology for constructing 3D facial meshes from speech cues captured in uncontrolled conditions. More particularly, we learned a statistical blendshape model by capturing 4D sequences of people uttering $500$ words selected from the Lip Reading Words (LRW) in-the-wild dataset. To align the words uttered from our participant with the words of the LRW, we proposed Deep Canonical Attentional Warping (DCAW), a novel method that simultaneously learns deep representations and an alignment path between two sequences. We thoroughly experimented with our proposed methods and showed the ability of a trained deep learning model on the create LRW-3D to generalise to different speakers and in uncontrolled conditions of speech.

For future work we intend to incorporate expression in our blendshapes such that accurate emotional speech can be obtained. In addition, we will test our model on languages different than English. On top of that, we will capture individuals talking on different languages to expand the generalisation capacity of our blendshapes.

\newpage
{\small
\bibliographystyle{ieee}
\bibliography{egbib}
}

\end{document}